# Feasibility of Identifying Factors Related to Alzheimer's Disease and Related Dementia in Real-World Data

Aokun Chen*[1], Qian Li*[1], Yu Huang*[1], Yongqiu Li[1], Yu-neng Chuang[2], Xia Hu[2], Serena Guo[3], Yonghui Wu[1], Yi Guo[1], Jiang Bian[1]

[1] Department of Health Outcomes and Biomedical Informatics, College of Medicine, University of Florida, 1889 Museum Rd, Suite 7000, Gainesville, FL 32610

[2] Department of Computer Science, George R. Brown School of Engineering, Rice University, 6100 Main St., Houston, TX 77005

[3] Department of Pharmaceutical Outcomes & Policy, College of Pharmacy, University of Florida, 1225 Center Drive, Gainesville, FL 32610

*Co-first, contributed equally

Corresponding authors:

Jiang Bian, PhD bianjiang@ufl.edu

Affiliation: Department of Health Outcomes & Biomedical Informatics, University of Florida

Address: 1889 Museum Rd #7002, PO Box 100147, Gainesville, FL 32610

Phone Number: (352) 273-8878

**ABSTRACT**

A comprehensive view of factors associated with AD/ADRD will significantly aid in studies to develop new treatments for AD/ADRD and identify high-risk populations and patients for prevention efforts. In our study, we summarized the risk factors for AD/ADRD by reviewing existing meta-analyses and review articles on risk and preventive factors for AD/ADRD. In total, we extracted 477 risk factors in 10 categories from 537 studies. We constructed an interactive knowledge map to disseminate our study results. Most of the risk factors are accessible from structured Electronic Health Records (EHRs), and clinical narratives show promise as information sources. However, evaluating genomic risk factors using RWD remains a challenge, as genetic testing for AD/ADRD is still not a common practice and is poorly documented in both structured and unstructured EHRs. Considering the constantly evolving research on AD/ADRD risk factors, literature mining via NLP methods offers a solution to automatically update our knowledge map.

**HIGHLIGHTS**

- We summarized the risk factors for AD/ADRD by reviewing existing meta-analyses and review articles on risk and preventive factors for AD /ADRD.
- Drawing from this literature review and identified AD/ADRD factors, we explored the accessibility of these risk and preventive factors in both structured and unstructured EHR data.
- We constructed an interactive knowledge map that can be used to aid in the design of future AD/ADRD studies that aim to leverage large collections of RWD to generate RWE.

## Introduction

Alzheimer's disease (AD) and AD-related dementias (AD/ADRD) are progressive neurodegenerative illnesses that cause memory loss and other cognitive impairments. In the United States, there are currently an estimated 6.7 million patients living with AD. This number is expected to double by 2060, reaching 13.8 million [1]. Despite being a national focus through initiatives like the National Alzheimer's Project Act and significant investments, there is still no effective treatment or preventive strategy. The few pharmaceutical treatments available primarily aim to alleviate symptoms, such as improving cognition, behavior, and global function; however, these improvements are modest at best [2]. Recently, new anti-amyloid antibody therapies, such as Aducanumab and Lecanemab, were approved by the U.S. Food and Drug Administration (FDA). These drugs have shown promising efficacy in clinical trials data [3,4], yet concerns about their real-world effectiveness and controversies have arisen [5–7], especially concerns on whether their limited efficacies reported in trials are clinically meaningful but also discussions on how to appropriately use these treatment in real-world clinical settings [8–10].

The complex mechanisms involved in the pathogenesis of AD/ADRD remain unclear. It is speculated that AD/ADRD results from a complicated interplay of brain changes associated with various age, genetic, environmental, and lifestyle factors. Although abundant literature exists on different factors, either as risks or protective elements, related to AD/ADRD from wet lab to population science studies, a comprehensive overview encompassing all these factors is lacking. Previous studies have often focused on subsets of these factors from specific types or categories, such as genetic or environmental influences [11–15]. A comprehensive view of factors associated with AD/ADRD will significantly aid in studies to develop new treatments for AD/ADRD and identify high-risk populations and patients for prevention efforts.

The wide adoption of electronic health records (EHRs) has made large collections of real-world data (RWD) [16,17] and detailed patient information available for research. This includes sociodemographic data, lab tests, medications, disease status, and treatment outcomes, offering unique opportunities to generate real-world evidence (RWE) that reflects the patients' treated in real-world settings. EHRs typically contain structured data, often coded (e.g., diagnoses coded in International Classification of Diseases). However, over 80% of information in EHRs is documented in free-text clinical narratives, such as physicians' notes and radiology reports [18]. These narratives contain detailed patient characteristics including important AD/ADRD risk factors, such as apolipoprotein E (APOE) and social determinants of health (SDoH), that are often not coded in structured EHR data. They also offer more fine-grained outcomes, such as scores from cognitive assessments like the Mini-Mental State Exam (MMSE) and Montreal Cognitive Assessment (MoCA), which help the determine severity of dementia. For example, in a previous study, we have assessed the documentation of cognitive tests and biomarkers for AD/ADRD in unstructured EHRs and developed natural language processing (NLP) pipelines to extract them into discrete values for downstream studies.

In our study, we summarized the risk factors for AD/ADRD by reviewing existing meta-analyses and review articles on risk and preventive factors for AD/ADRD. Compared with the previous studies [19–24], our work provides a updated review of recent publications on AD/ADRD-related risk factors. Drawing from this literature review and identified AD/ADRD factors, we explored the accessibility of these risk and preventive factors in both structured and unstructured EHR data. To disseminate our study results, we also constructed an interactive knowledge map that can be used to aid in the design of future AD/ADRD studies that aim to leverage large collections of RWD to generate RWE.

**Methods**

Our study consisted of four steps: literature screening, full-text extraction, knowledge graph generation, and accessibility assessment, as depicted in **Figure 1**. We first searched and screened titles and abstracts from relevant literature databases using a predefined set of keywords. We then performed full text extraction on the selected literature on risk factors for AD/ADRD. With the extracted AD/ADRD risk factors, we constructed a knowledge graph and visualized these factors and their interrelationships. Finally, we examined the accessibility of these AD/ADRD risk factors based on a RWD dataset from the University of Florida Health (UF Health) Integrated Data Repository (IDR), with a specific focus on those factors not readily available in structured EHRs, but in unstructured clinical narratives.

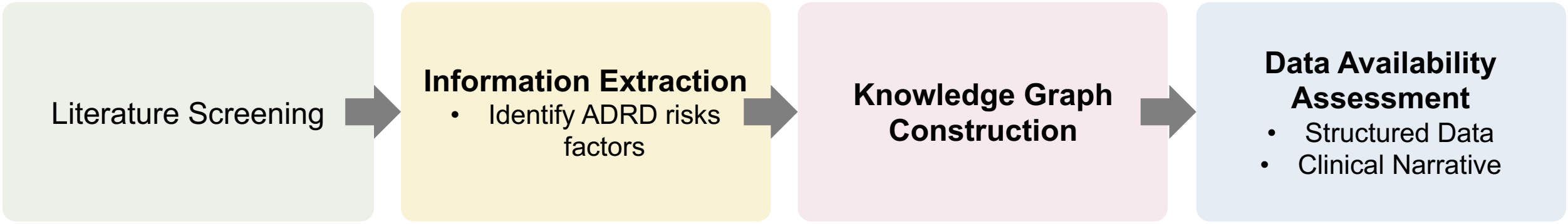

**Figure 1.** The overall workflow of the study.

<u>Literature search strategy and screening</u>

Our literature review process adhered to the Preferred Reporting Items for Systematic Reviews and Meta-Analyses (PRISMA) guidelines. To thoroughly identify known risk or protective factors related to AD/ADRD, we conducted searches in three electronic bibliographic databases—PubMed, Cochrane, and Embase—focusing on reviews, systematic reviews, and meta-analysis articles published in the most recent decade. We included two sets of keywords: (1) those related to AD/ADRD (e.g., “Alzheimer’s disease,” “Alzheimer’s and related dementias,” “ADRD,” “Lewy body,” “vascular dementia,” “frontotemporal dementia,” and “dementias”) and (2) those related to risk or protective factors (e.g., "risk factor," and "prevention"). We excluded case reports, non-human studies, studies of factors not directly associated with dementia, and non-English literature from the search results. We conducted two rounds of screening: title/abstract screening and full-text screening. In each round, at least two reviewers independently reviewed the materials, with any conflicts resolved by a third reviewer.

Full-text risk factor extraction

We developed an information extraction form to document the risk factors and their relationships with the outcomes of interest reported in each article. The relationship should include three components: the factor, the outcome, and the effect, formulating a triple statement in the format of a subject-predicate-object expression. For example, one article reported, "*A meta-analysis of studies in normal individuals detected that higher adherence to the MedDiet was associated with lower risk for Alzheimer's disease (AD) (HR = 0.64; 95 % CI:0.46, 0.89)*" [25]. Here, the factor is "*higher adherence to the MedDiet*", the outcome is "*risk for Alzheimer's disease*", the effect is "*decreased risk significantly,*" and the triple is "*higher adherence to the MedDiet*"- "*decreased (significantly)*" - "*risk for Alzheimer's disease.*" After extraction, we developed standardized categories for each component to summarize the factors, outcomes, and the effects of the factors on the outcomes (**Table 1**).

**Table 1**. Standardized categories for the risk factors, outcomes, and the effects of the factors on the outcomes.

| Outcome | Factor Category | Effect |
|---|---|---|
| • Risk of all-cause dementia<br>• Risk of Alzheimer's Disease<br>• Risk of Vascular Dementia<br>• Risk of Frontotemporal Dementia<br>• Risk of Lewy Body Dementia | • Genomic<br>• Condition<br>• Lifestyle<br>• Biomarker<br>• Medication<br>• Procedure<br>• Family history<br>• Environment<br>• Social economic status<br>• Demographic | • Increase risk significantly<br>• Decrease risk significantly<br>• Not significant (not enough evidence / no association)<br>• Inconsistent evidence |

Knowledge graph construction

With the triples extracted from literature, we created a knowledge graph on ADRD risk factors using Neo4j. This knowledge map encompasses the risk factors, associated outcomes, source literature, categories, and their effects on AD/ADRD risk. We have made this knowledge graph available in an open-access domain to facilitate the exploration of our review results by other researchers.

Assessment of data availability in electronic health records (EHRs)

We evaluated the data availability of the extracted AD/ADRD-related factors based on the EHRs of an AD/ADRD patient cohort retrieved from the University of Florida (UF) Health Integrated Data Repository (IDR). UF IDR is a clinical data warehouse of UF Health clinical and research enterprises. It consolidates information from various clinical and administrative information systems, including the Epic EHR system, into the IDR data warehouse. The IDR contains more than 2 billion observational facts pertaining to more than 2 million patients. Our assessment consisted of two parts, reflecting the nature of EHR data, (1) structured and coded EHRs, and (2) unstructured clinical narratives such as physicians' progress notes and various reports.

For structured EHR data, we considered two layers of availability: (1) whether the data model can capture the information, and (2) whether the information is actually present in the data. We considered two widely used Common Data Models (CDMs): the National Patient-Centered Clinical Research Network (PCORnet) CDM, and the Observational Medical Outcomes Partnership (OMOP) CDM, as UF Health is part of the OneFlorida+ Clinical Research Consortium contributing to the national PCORnet [26]. According to the two CDMs, we determine that the data model can capture the factors if the data fields are directly available (e.g., age, gender) or if the factors could be queried using standardized codes (e.g., International Classification of Diseases [ICD] codes for conditions and diseases). For each factor that can be formalized into a medical coding system (e.g., ICD codes, Current Procedural Terminology [CPT] codes, Healthcare Common Procedure Coding System [HCPCS] codes, Logical Observation Identifiers Names and Codes [LOINC], RxNorm codes, or National Drug Code [NDC]), we map the factors to the coding system in the data following PCORnet and OMOP CDMs. When then query UF Health IDR to determine whether the risk factors are actually being documented in EHRs.

Some risk factors, such as cognitive tests like MMSE and MoCA, may only be documented in clinical narratives. To assess the availability of each risk factor in unstructured EHR data, we first developed a set of keywords considering various variations, including known abbreviations and synonyms, of how the risk factors would be documented in clinical notes, using a snowballing approach. Using a set of seed keywords (e.g., "MMSE," "Mini-Mental State Exam", etc), we searched the keyword patterns in the clinical narratives, reviewed a random sample (n=20) of the hits, and determined whether each sample contain the information about the risk factor of interest. We calculated the precision (i.e., number of notes that does contain the risk factor over the total number of sampled notes) and eliminated keyword patterns that have high false positive rates. During this process, we also added new keyword patterns (e.g., a new synonym that we did not capture). We iterated these steps until we did not find any new keyword patterns for each factor.

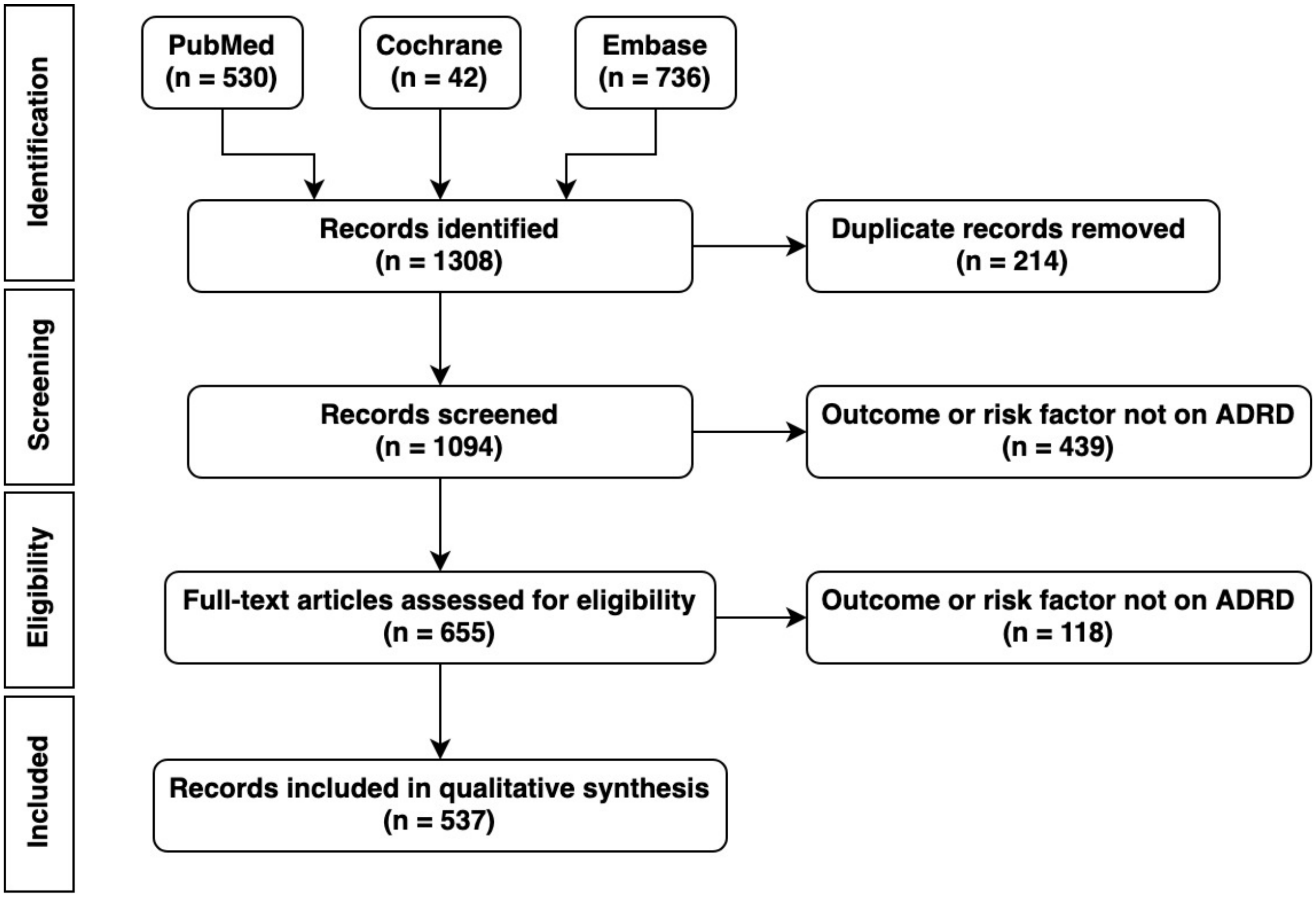

**Figure 2**. Our overall literature search and review procedure according to the Preferred Reporting Items for Systematic Reviews and Meta-Analyses (PRISMA) flow diagram.

**Results**

A total of 1,308 studies from the last decade were identified from the literature databases. After de-duplication, 1,094 studies remained. Error! Reference source not found. shows our review procedure in a PRISMA flow diagram. The search yielded 537 distinct studies focused on identifying ADRD-related risk factors.

**AD/ADRD risk factors in RWD literature**

In total, we extracted 477 risk factors in 10 categories from the identified studies. Detailed statistics on these risk factors is shown in **Table 2**.

**Table 2**. Summary statistics of AD/ADRD risk factors extracted from literature.

| Risk factor category | Number of unique risk factors | Number of unique articles |
|---|---:|---:|
| Genomic | 88 | 55 |
| Disease/Condition | 172 | 88 |
| Lifestyle | 73 | 42 |
| Biomarker | 40 | 24 |
| Medication | 71 | 48 |
| Procedure | 2 | 4 |
| Family history | 1 | 1 |
| Environment | 6 | 5 |
| Socioeconomic status | 16 | 13 |
| Demographic | 8 | 10 |

Genomic-related risk factors reported in the literature for AD/ADRD typically involve gene mutations and alleles that could affect the incidence and outcome of the disease. We identified 88 unique genomic-related AD/ADRD risk factors from 55 studies [27–80]. Among these, 81 were reported to be related to the risk of AD, 2 on vascular dementia, and 5 on other types of dementia. Out of these studies, 40 reported genomic risk factors that increased the risk of AD/ADRD [27,28,32–36,39–42,44,46–50,53,54,56,58–65,67,72–81], 17 reported non-significant effect [29,37,38,51,55,60,62,70,71], 4 indicated genomic-related risk factors that lower AD/ADRD risk [57,65,66,68], and 1 reported inconsistent results [31]. The most frequently reported genomic-related risk factors is the apolipoprotein E (APOE) gene's alleles (i.e., APOE, APOE e2, APOE e2/3, APOE e3, APOE e3/4, APOE e4, APOE e4/4), mentioned in a total of 23 mentions. Among the APOE mutations, APOE e3 was consistently reported to lower the risk, and APOE e4 to increase the risk, across various studies, while the effects for the other APOE alleles were mixed.

Disease or condition risk factors refer to other health issues of AD/ADRD patients. We identified 172 unique risk factors from 88 literature. [34,36,69,82–152] Among these, 61 were related to AD, 14 to vascular dementia, and 97 to other types of dementia. Of the 88 studies, 76 reported various diseases and conditions that increased the risk of AD/ADRD [34,36,69,82–85,87–96,98,100–105,107–117,120,121,123–137,139–144,146–148,151,153–161], 13 reported a non-significant effect [91,95,105,109,110,128,145–147,149,150,152,160], 3 studies indicated that certain conditions to lower the risk [114,146,147], and 9 reported inconsistent results [14,86,97,99,106,118,122,138,158]. The five most frequently analyzed conditions included depression (mentioned in 31 studies), diabetes (i.e., type 1 and type 2 diabetes, in 27 studies), traumatic brain injury (14 studies), hypertension (14 studies), and anxiety (13 studies). In the surveyed literature, depression (with 29 studies reported an increased risk of AD/ADRD vs. 2 inconsistent), diabetes (25 increase vs. 1 inconsistent vs. 1 decrease), hypertension (12 increase vs. 2 inconsistent), and anxiety (9 increase vs. 4 inconsistent) were reported to increase the risk of AD/ADRD. Traumatic brain injury showed mixed outcomes, with 7 studies reporting an increased risk of AD/ADRD, 6 inconsistent, and 1 a decrease.

Lifestyle risk factors for AD/ADRD include factors such as physical activity, diet, and substance use. We identified 73 unique lifestyle-related risk factors from 42 publications [34,36,69,91,95,109,110,114,120,122,123,158,162–190]. Among these, 43 were related to AD, 6 to vascular dementia, and 53 to other types of dementia. Of the 42 studies, 14 reported lifestyle risk factors increased the risk of AD/ADRD [69,109,110,120,123,158,162,164,178,181,183,184,189,190], 24 indicated a decrease in risk [34,36,69,109,110,114,158,162,163,167,168,170,172–180,182,187,190], and 5 reported inconsistent effects [91,95,106,122,188]. The five most frequently mentioned lifestyle risk factors in the literature were physical activity (mentioned in 23 studies), smoking (19 studies), alcohol (11 studies), Mediterranean diet (9 studies), and diet in general (7 studies). Regarding physical activity, there was a general consensus that an increasing in activity (1 inconsistent vs. 15 decrease) decreased the risk of AD/ADRD, while the impact of inactivity was less clear (3 increase vs. 4 inconsistent). Most studies suggested that smoking increased the risk of AD/ADRD (13 increase vs. 4 inconsistent vs. 2 decrease). Moderate alcohol consumption was generally believed to reduce the risk of AD/ADRD (increase vs. inconsistent vs. decrease = 3:2:6). As for diet, a healthy diet, such as Mediterranean diet, was considered to decrease the risk of AD/ADRD (3 inconsistent vs. 13 decrease), while diets high in saturated fat were reported to increase the AD/ADRD risk in one literature.

Biomarkers are objective measures that can be used to indicate a patient's medical state accurately and reproducibly, and can be and often used to assess the risk of the patients [191]. We identified 40 unique biomarkers from 24 studies [110,114,155,192–210]. Among these, 39 were found related to AD, 7 to vascular dementia, and 23 to other types of dementia. Of the 24 studies, 26 biomarkers were found to indicate an increased risk of AD/ADRD [109,110,114,192–200,204–210], 2 were associated with a decreased risk [155,203], and 2 showed inconsistent results [122,196]. The seven most studied biomarkers included tau, cholesterol, homocysteine, bone mineral density, magnesium, white matter hyperintensities, and vitamin D level. Most literature reported an increase in tau levels (4 increase vs. 1 inconsistent), homocysteine levels (4 increase), vitamin D deficiency (3 increase), low bone mineral density (2 increase vs. 1 inconsistent), low magnesium levels (2 increase vs. 1 inconsistent), and white matter hyperintensities (2 increase vs. 1 inconsistent) are associated with an increased risk of AD/ADRD. The impact of cholesterol levels was found to be inconsistent in 4 studies.

Medication-related factors, including dietary supplements, refer to the use of medications that could affect the risk of AD/ADRD. We identified 71 unique medication-related risk factors from 48 publications [34,36,69,109,110,114,158,211–249]. Specifically, 62 factors were associated with AD, 3 with vascular dementia, and 41 with other types of dementia. Among these, 3 medications were found to increase the risk of AD/ADRD [69,158,231,250], 45 medications were associated with decreased risk [34,36,69,109,110,114,158,212,217,218,220–222,225,227–230,232–236,238–

240,242,244,245,247], 21 showed insignificant results [109,110,158,213–216,219,223,226,228,232,235,237,241,243,248,249,251], and 2 yielded inconsistent results [211,224,246]. The most frequently mentioned medications were statins, antihypertensive medications, Omega-3 Fatty Acids supplement, vitamin E supplements, hormone therapy, and memantine—an antagonist of the N-Methyl-D-Aspartate (NMDA)-receptor used to slow the neurotoxicity that thought to be involved in AD and other neurodegenerative diseases. Most studies reported statins (9 decrease vs. 5 inconsistent), antihypertensive medications (9 decrease vs. 3 inconsistent), vitamin E (3 decrease vs. 2 inconsistent), and memantine (4 decrease) as reducing AD/ADRD risk. The effect of Omega-3 Fatty Acids (1 decrease vs. 4 inconsistent) and hormone therapy (3 increase vs. 3 inconsistent) on AD/ADRD risk were found to be inconsistent.

Procedures in this context refer to medical procedures or non-pharmaceutical interventions that could potentially affect the risk of AD/ADRD. We identified 2 unique procedure-related risk factors from 4 studies [252–255], where neither cognitive training exercises or anesthesia had a significant effect on AD/ADRD risk.

The risk factor of family history pertains to ancestral health patterns that could affect the risk of AD/ADRD. We identified one specific family history risk factor from our review: a family history of Parkinson's disease [256], however its effect on AD risk was not significant. This also speaks to the fact that family history information is poorly documented in EHRs, leading to limited studies that examined how family history affect AD/ADRD risk.

Environment risk factors refer to aspects of ones' surroundings (e.g., natural and built environments) that could influence the risk of AD/ADRD. We identified 6 unique environmental factors from 5 studies [109,110,257–261]. Among these, 6 factors were reported in at least 1 publication to increase the risk of AD/ADRD [257–260], one factor was found to decrease AD/ADRD risk [261], and another was reported to have an insignificant effect [109,110]. The three most frequently mentioned environmental risk factors were electromagnetic fields, pesticides, and air pollution, and all of them, electromagnetic fields (4 increase), air pollution (3 increase vs. 1 decrease), and pesticide (3 increase vs. 1 insignificant) were reported to increase the risk of AD/ADRD.

Socioeconomic status factors represent the aspects of social and economic standing that influence an individual's AD/ADRD risk. We identified 16 unique socioeconomic status risk factors from 13 studies [37,106,109,110,114,116,117,122,158,262–266]. Of these, 6 were reported to affect the risk of AD, while the rest were associated with dementia in general. The most frequently mentioned socioeconomic factor was education, cited in 13 studies. Most of these studies indicated that a higher level of education is associated with a decreased risk of AD/ADRD (15 decrease vs. 5 inconsistent).

Demographic risk factors pertain to the patients' demographic characteristics that are associated with their AD/ADRD risk. We identified 8 unique demographic risk factors from 10 studies [15,34,36,109,110,122,264,267–269]. Among these, 2 risk factor were related to AD, 1 to frontotemporal dementia, and 5 others were related to the risk of dementia in general. The three most frequently mentioned risk factors are age, sex, and bilingualism. All studies addressing age (increase = 5) reported being elderly increases the risk of AD/ADRD. The effect of sex (3 inconsistent vs. 1 increase) and bilingualism (2 inconsistent) on AD/ADRD risk were found to be inconsistent.

<u>**Availability of risk factors in real-world data (RWD)**</u>

In addition to surveying the risk factors for AD/ADRD, we also evaluated their accessibility in RWD, especially EHRs. We emphasized risk factors that are not readily available in structured EHRs, but may exist in unstructured clinical narratives, spanning the 10 categories: genomics, condition or disease, lifestyle, biomarker, medication, procedure, family history, environment, socioeconomic status, and demographics. To assess the accessibility of these risk factors, we extracted both structured EHRs and unstructured clinical notes from AD/ADRD patients covering the period from 2012 to 2020 from UF Health IDR. The demographic information of our AD/ADRD cohort can be found in **Table 3**.

**Table 3.** Demographics of AD/ADRD cohort from University of Florida (UF) Health Integrated Data Repository (IDR).

| | **AD/ADRD patients from UF Health IDR (N=48,912)** |
|---|---:|
| Age | |
| Mean (SD) | 67.94 (21.08) yrs |
| Sex | |
| Male (%) | 23,062 (47.15%) |
| Race | |
| White (%) | 34,252 (70.03%) |
| Black (%) | 11,120 (22.74%) |
| Other (%) | 2,693 (5.50%) |
| Unknown (%) | 846 (1.73%) |
| Ethnicity | |
| Hispanic (%) | 1,851 (3.78%) |
| Non-Hispanic (%) | 45,462 (92.95%) |
| Unknown (%) | 1,599 (3.27%) |

Among structured EHRs, we identified 250 out of 477 (55.9%) risk factors. Breaking this down by category, we found 171 condition or disease factors (99.42% out of the total 172), 23 biomarkers (57.50% out of 40), 47 medications (66.20% out of 71), 1 procedure-related factor (50% out of 2), and 7 demographic factors (87.50% out of 8) within the structured EHR. However, we were unable to identify any risk factors in the categories of genomics, lifestyle, family history, environment, and socioeconomic status.

We further examined the availability of these risk factors within unstructured clinical narratives. As shown in **Table 4**, we identified 41 instances of keywords related to AD/ADRD risk factors within the clinical narratives of our AD/ADRD cohort from the UF Health IDR. The top nine most frequently mentioned risk factors among the AD/ADRD patients in clinical narratives included alcohol (96.64%), smoking (96.62%), education (94.24%), activity (i.e., outdoor activity, indoor activity, lack of activity, etc., 92.56%), occupation (88.35%), diet (87.53%), vitamin (75.23%), exercise (e.g., swim, walk, etc. 74.96%), and environment (60.69%). Broken down by the 10 categories, we identified 5 genomic risk factors (5.68% of the 88 genomic risk factors), 14 lifestyle risk factors (19.18% of 73), 3 biomarkers (7.5% of 40), 10 medications (14.08% of 71 ), 5 environment risk factors (83.33% of 6), 3 socioeconomic status factors (18.75% of the 16), and 1 demographic factor (18.75% of 8) that are available in unstructured EHRs. The summary statistics of the risk factors mentioned in clinical narratives from the UF Health AD/ADRD cohort are summarized in **Table 4**.

**Table 4**. AD/ADRD risk factors identified in clinical narratives from the UF Health AD/ADRD cohort.

| **Risk factor** | **Unique # of clinical notes with relevant keywords (N = 17,233,317)** | **Unique # of patients with relevant keywords (N=48,912)** |
|---|---|---|
| **Genomic** | | |
| APOE | 8,429 (0.05%) | 1,174 (2.45%) |
| BIN1 | 2 (< 0.01%) | 1 (< 0.01%) |
| Gene (non-specific) | 60,192 (0.35%) | 6,239 (13.0%) |
| Polymorphism | 436 (0.0%) | 92 (0.19%) |
| **Lifestyle** | | |
| Activity (non-specific) | 2,514,341 (14.59%) | 44,414 (92.56%) |
| Agreeableness | 16 (0.0%) | 13 (0.03%) |
| Alcohol | 3,064,941 (17.78%) | 46,372 (96.64%) |
| Coffee | 103331 (0.6%) | 16,664 (34.73%) |
| Diet | 2,035,016 (11.81%) | 41,999 (87.53%) |
| Exercise (non-specific) | 841,177 (4.88%) | 35,970 (74.96%) |
| Extraversion | 7 (< 0.01%) | 6 (0.01%) |
| Fish consumption | 236,192 (1.37%) | 16,292 (33.95%) |
| Marital status | 732,762 (4.25%) | 28,598 (59.60%) |
| Mastication | 47,322 (0.27%) | 12,253 (25.54%) |
| Neuroticism | 4 (< 0.01%) | 3 (0.01%) |
| Openness | 379 (< 0.01%) | 189 (0.39%) |
| Smoking | 3,195,170 (18.54%) | 46,360 (96.62%) |
| Yoga | 20,399 (0.12%) | 5,526 (11.52%) |
| **Biomarker** | | |
| Estrogen | 25,593 (0.15%) | 4,961 (10.34%) |
| Estrone | 34 (< 0.01%) | 14 (0.03%) |
| Hormones | 13,854 (0.08%) | 4,730 (9.86%) |
| **Medication** | | |
| Antioxidant | 1,377 (0.01%) | 189 (0.39%) |
| Cannabinoids | 5,334 (0.03%) | 1,159 (2.42%) |
| DHA | 6,711 (0.04%) | 516 (1.08%) |
| Diuretics | 81,871 (0.47%) | 10,213 (21.28%) |
| Folate | 294,293 (1.71%) | 25,180 (52.48%) |
| Lycopene | 686 (< 0.01%) | 127 (0.26%) |
| Omega | 219,127 (1.27%) | 7,184 (14.97%) |
| Polyphenols | 668 (< 0.01%) | 123 (0.26%) |
| Vitamin | 2,030,498 (11.78%) | 36,100 (75.23%) |
| Zinc | 165,757 (0.96%) | 8,169 (17.02%) |
| **Environment** | | |
| Aluminum | 257,650 (1.49%) | 13,160 (27.43%) |
| Electromagnetic | 563 (< 0.01%) | 272 (0.57%) |
| Environment (non-specific) | 287,097 (1.67%) | 29,120 (60.69%) |
| Pesticides | 1,143 (0.01%) | 356 (0.74%) |
| Solvents | 449 (< 0.01%) | 118 (0.25%) |
| **Socioeconomic status** | | |
| Education | 2,759,829 (16.01%) | 45,219 (94.24%) |
| Illiteracy | 678 (< 0.01%) | 122 (0.25%) |
| Occupation | 1,246,606 (7.23%) | 42,394 (88.35%) |
| **Demographic** | | |
| Bilingualism | 4 (< 0.01%) | 2 (< 0.01%) |

## A knowledge graph of AD/ADRD risk factors in RWD

With the risk factors and their relations to outcomes identified from the reviewed studies, we built an interactive knowledge graph using Neo4j—a graph database. This knowledge graph visualizes the risk factors and their related mentions from the literature. The knowledge graph and instructions for setting up the Neo4j can be accessed at https://github.com/uf-hobi-informatics-lab/ADRD_risk_factor_knowledge_map.

With the established triple statements among risk factors and AD/ADRD outcomes mined from the literature, the knowledge embedded in the graph is poised to facilitate the design of new studies on AD/ADRD using RWD. Below, we demonstrate an example of using the knowledge graph to identify ADRD-related risk factors, potential outcomes, and the interconnections between different risk factors. This use case illustrates how researchers can gain a global view of existing AD/ADRD risk factors through the knowledge map and tailor the scope of their investigation accordingly.

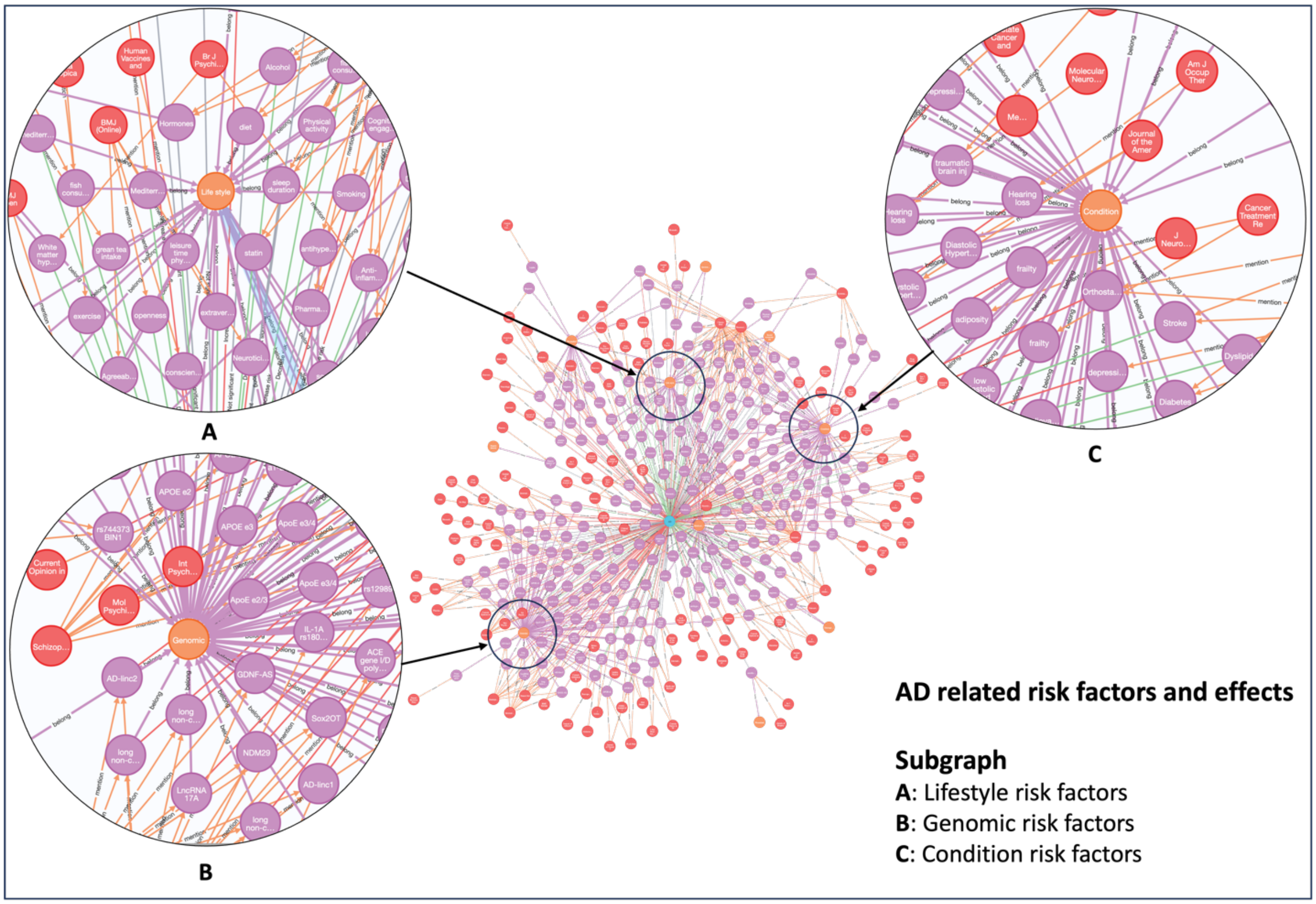

**Figure 3**. An overview of the knowledge graph for AD-related risk factors.

**Figure 3** shows a global view of the knowledge graph with AD-related factors, where the purple dots are the risk factors, while the red dots indicate related mentions identified from the literature. In total, there are 246 AD-related risk factors from 251 related literature mentions. We can further navigate the knowledge graph to focus on demographic and environmental risk factors related to AD. From **Figure 4-a**, we can see that the risk of AD increases as the age increases reported in two studies: Hersi et al. [270] and Navipour et al. [124]. **Figure 4-b** highlights findings from two studies, Angehrn et al. [257] and Rawat et al. [40], suggesting that environmental exposure to electronic fields may contribute to the development of AD. Additionally, two other studies, Nordestgaard et al. [116] and Gunnarsson et al. [271], concluded that pesticide exposure does not have a significant effect on AD.

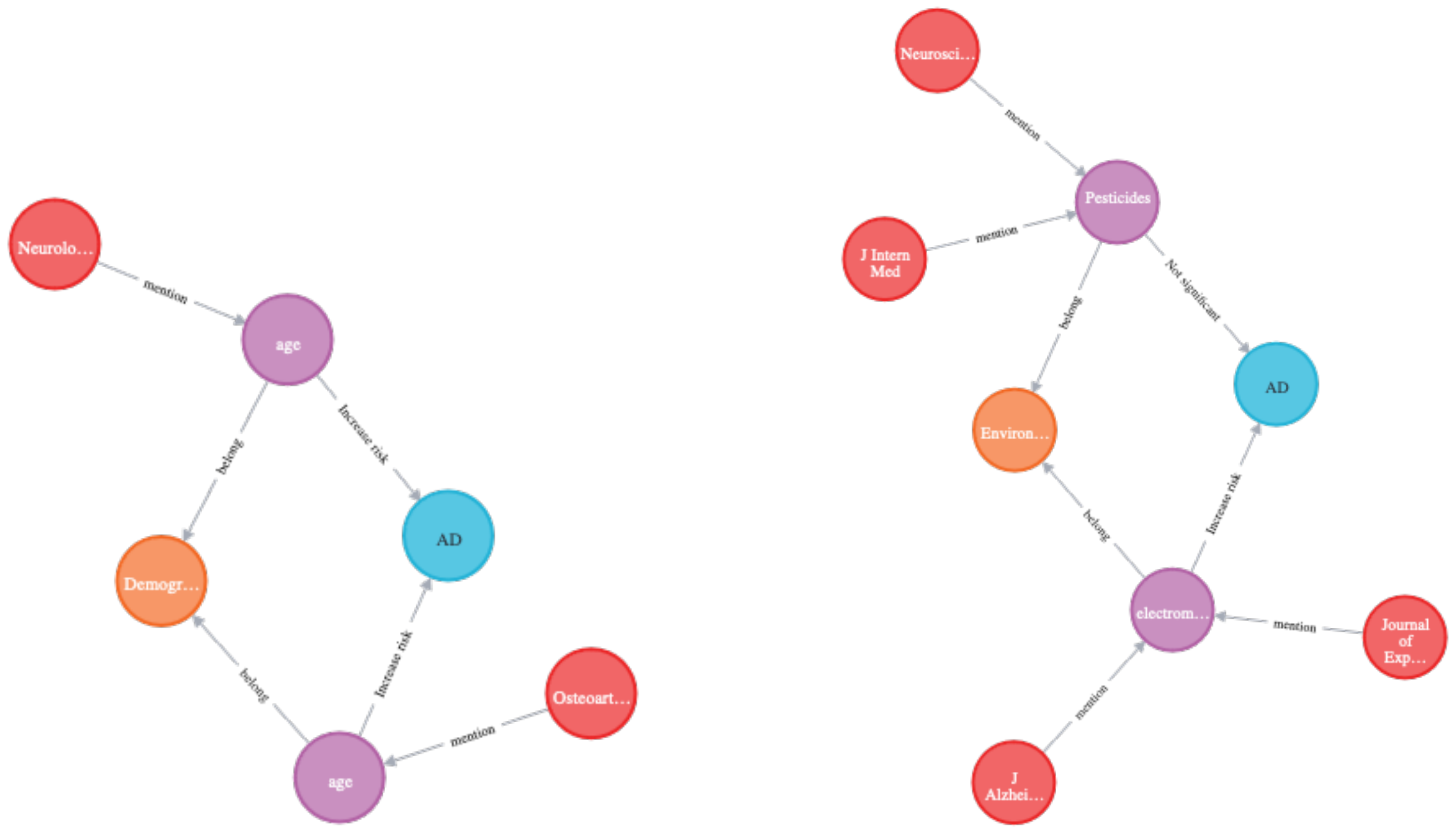

(a) (b)

**Figure 4**. Demographic and environmental risk factors for AD: (a) illustrates that age is a risk factor of AD, and (b) illustrates that exposure to pesticide does not have significant effect on AD.

## Discussion

Through a systematic search of literature that focused on AD/ADRD risk factors from the last decade, we extracted 477 risk factors for AD/ADRD in 10 categories, including genomic (n=88), condition or disease (n=172), lifestyle (n=73), biomarker (n=40), medication (n=71), procedure (n=2), family history (n=1), environment (n=6), socioeconomic status (n=16), and demographics (n=8), from 537 studies that met our criteria. We further analyzed the accessibility of these risk factors in both structured and unstructured (i.e., clinical narratives) EHR data using an AD/ADRD cohort from UF Health IDR. We found that 55.9% of the risk factors, primarily genomic, lifestyle, environment, socioeconomic status factors, were not available in structured EHRs, but do exist in unstructured clinical narratives.

Among the top five mentioned AD/ADRD risk factors, condition or disease-related risk factors were the most frequently studied, with the majority indicating an increased risk of AD/ADRD. Genomic factors ranked second in terms of frequency, with most also reported to elevate AD/ADRD risk. Lifestyle factors were the third most commonly studies category, and these were also predominantly found to increase AD/ADRD risk. Medication factors, the fourth most researched category, were generally found to decrease the risk of AD/ADRD. Biomarker factors, ranking fifth, were mostly reported to associate with an increased risk of AD/ADRD, understandably as these biomarkers are developed to identify or measure AD/ADRD.

Regarding the availability of the top five AD/ADRD risk factors, unsurprisingly a significant portion of condition or disease-related risk factors were accessible from structured EHRs (171 out of 172, 99.42%). This high accessibility likely contributes to their prevalent mention in the literature. In contrast, the availability of genomic risk factors in both structured and unstructured EHR data is limited, despite their frequent mention in the literature; nevertheless, these mentions are mostly focused on the APOE gene—a widely known genetic risk factor for AD. Only 4 genomic risk factors (4.55% out of a total of 88) were only found in clinical narratives of our AD/ADRD cohort from UF Health. The reason is potentially two-fold: (1) genetic testing for AD/ADRD has only become available in the last decade and not yet a routine clinical practice, and (2) genetic tests are often conducted by external labs, with results typically provided in PDF report format rather than being stored discretely in the EHR system. At UF Health, we have only recently implemented the Genomics module in Epic, which is capable of ingesting genetic results from external labs through a standardized interface, thereby making them available in a discrete format. In terms of lifestyle factors, none were retrievable from the structured EHRs. However, from the clinical narratives, we retrieved 14 lifestyle risk factors, accounting for 19.18% out of the total 73, where the most frequently mentioned lifestyle factors include alcohol consumption, smoking, physical activity, and diet. Regarding medication risk factors, 47 (66.20% out of a total of 71) were available from the structured EHRs. These 47 risk factors predominantly involved prescription medicines, making it the second most common category of risk factors sourced from structured EHRs. Additionally, from the clinical narrative, we extracted 10 medication risk factors involving over the counter (OTC) medications, increasing the overall availability of medication risk factors to 80.28%. Among the biomarkers, 23 (57.5% out of a total of 40) were available from structured EHRs. Incorporating the 3 that can be extracted from clinical narratives increased the availability of biomarker risk factor to 65%.

From our observations, we noted a clear trend in the availability of AD/ADRD risk factors within structured EHRs, correlating with their prevalence in literature, except for genomic risk factors. This observation inspires two avenues of future research. First, there is a need for NLP tools to extract AD/ADRD risk factors, especially in the areas of lifestyle, family history, environment, and socioeconomic categories. These AD/ADRD risk factor categories have been under-researched, likely due to limited data availability. Our results indicate that clinical narratives are a promising information source within EHRs. An NLP-based extraction system could greatly enhance data availability in these categories, thereby facilitating research on corresponding risk factors. Second, the integration of other data and information sources with the EHR data is necessary to capture a complete picture of the patient disease development process. For example, genomic factors are the second most researched category, however, they are poorly populated in EHRs. Most current AD/ADRD genomics research data originate from cohorts established by various National Institute on Aging (NIA)-funded consortiums, centers, and repositories [272]. However, these cohorts often lack comprehensive phenotypic information and other critical clinical and socio-environmental factors for modeling AD/ADRD, which are rich in RWD like EHRs. The ability to link and integrate RWD with other data sources, encompassing a broad range of information domains, is critical for future AD/ADRD research. Indeed, the NIA has established the NIA Data LINKAGE Program (LINKAGE) in 2021, aiming to connect NIA-funded study data with other datasets, particularly RWD such as the Medicare claims data from the Centers for Medicare & Medicaid Services (CMS). It also provides a cloud-based Enclave environment to facilitate additional linkages and analyses of these integrated datasets [273].

The knowledge graph needs to be continuously updated, as research on AD/ADRD continues to evolve. However, manually extracting AD/ADRD risk factors from published studies through a systematic scoping review is time-consuming. Literature mining via NLP methods offers an automated way to extract risk factors and relationships from relevant literature, facilitating the discovery of potential connections among AD/ADRD risk factors and outcomes. For example, our previous study [274] demonstrated the effectiveness of employing entity recognition and relation extraction methods to automatically construct a knowledge graph by mining the abstracts of relevant literature. As an illustration, we conducted a search to investigate the relationship between brain trauma and dementia. The results of this search are presented in **Figure 5**. As depicted in the figure, literature reference #632 [275] suggests that traumatic brain injuries, including mild traumatic brain injuries, are likely to contribute to the development of dementia, with trauma potentially acting as a risk factor for AD-related dementia.

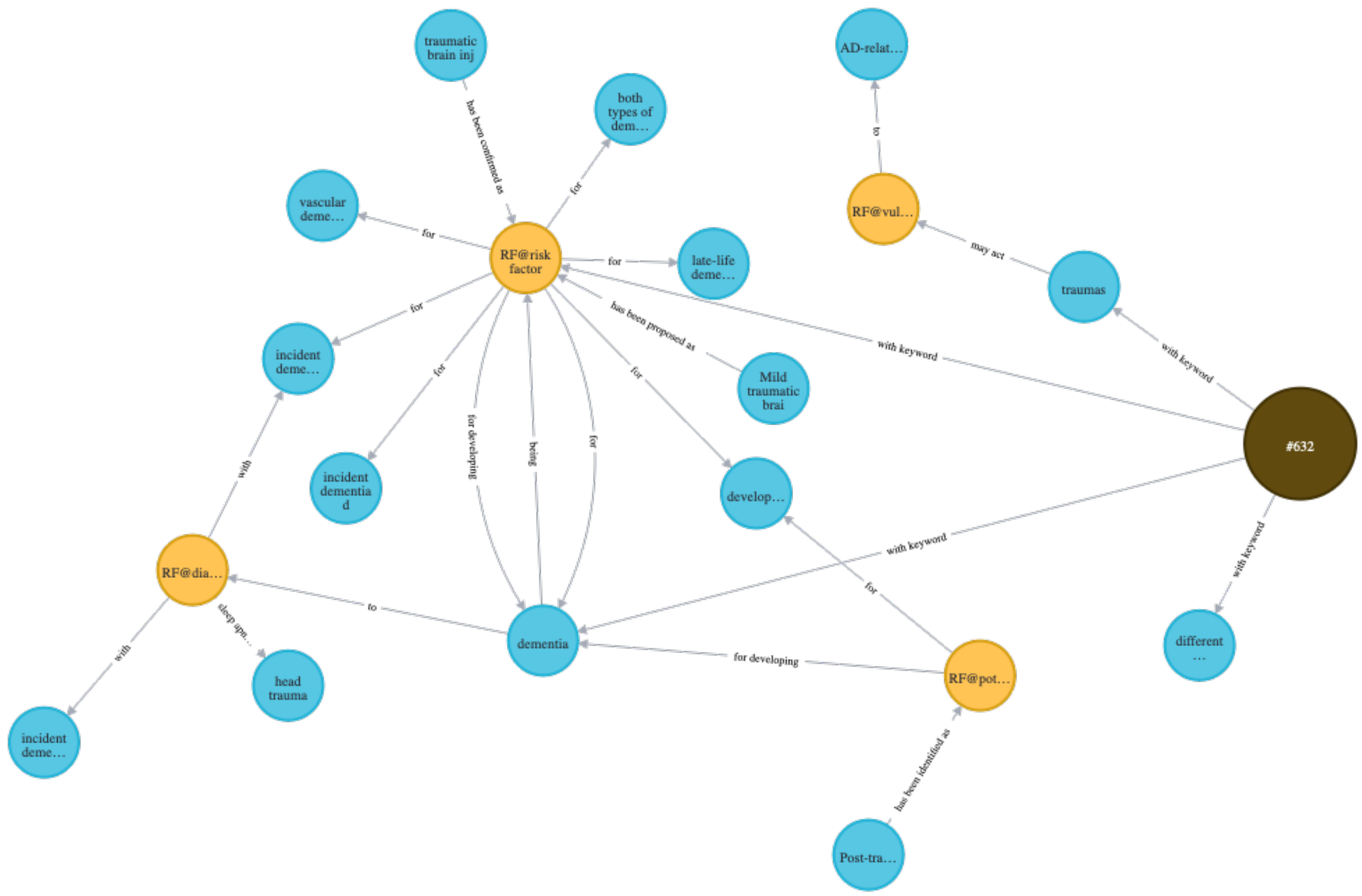

**Figure 5**. Relation between brain trauma and dementia mined from literature via natural language process methods.

Our study had limitations. First, we excluded studies not written in English and only focused on reviews, systematic reviews, and meta-analysis articles that were peer-reviewed in the last decade. Consequently, our survey may have missed the most recently published literature. Second, we evaluated the availability of AD/ADRD risk factors based on EHR data from a single health system—UF Health IDR. As data models and documentation patterns vary among different EHR systems and clinical practices, the availability of AD/ADRD risk factors might differ accordingly. Lastly, our evaluation of AD/ADRD risk factors relied on a keyword-based method. This approach may have overlooked some risk factors, as it is impractical to include every possible keywords.

## Conclusion

In this study, we conducted a literature review on AD/ADRD risk factors, identifying 10 categories encompassing 477 risk factors. Existing studies predominantly focus on condition or disease-related risk factors, medication risk factors, lifestyle risk factors, and genomic risk factors.

Most of the risk factors (including condition, medication, condition, biomarker, and procedure) are accessible from structured EHRs. For those risk factors not accessible from structured EHRs, clinical narratives show promise as information sources, particularly for lifestyle, environmental, and socioeconomic status factors. These narratives also supplement the data on OTC medications and dietary supplements within the category of medication risk factors. However, evaluating genomic risk factors using RWD remains a challenge, as genetic testing for AD/ADRD is still not a common practice as well as being poorly documented in both structured and unstructured EHRs. Our study provides valuable insights and interactive materials to researchers regarding AD/ADRD-related risk factors in RWD and highlights gaps in the field.

**Acknowledgement:**

This study was funded by National Institute on Aging grant R01AG080991.